%% file: main.tex
\documentclass{article}

\PassOptionsToPackage{table}{xcolor}
\usepackage[preprint]{neurips_2026}

\usepackage[utf8]{inputenc}      % allow utf-8 input
\usepackage[T1]{fontenc}         % use 8-bit T1 fonts
\usepackage{hyperref}            % hyperlinks
\usepackage{url}                 % simple URL typesetting
\usepackage{booktabs}            % professional-quality tables
\usepackage{amsfonts}            % blackboard math symbols
\usepackage{amsmath}             % advanced math typesetting
\usepackage{amssymb}             % more math symbols
\usepackage{nicefrac}            % compact symbols for 1/2, etc.
\usepackage{microtype}           % microtypography
\usepackage{xcolor}              % colors
\usepackage{graphicx}            % include figures
\usepackage{multirow}            % multirow table cells
\usepackage{array}               % extended array environment

\definecolor{fpgray}{HTML}{F3F4F6}
\definecolor{oursblue}{HTML}{E8F2FF}
\definecolor{qbitsgreen}{HTML}{EEF8F0}
\newcommand{\bestq}[1]{\textbf{#1}}
\newcommand{\oursmethod}{\textbf{RPFQ}}
\newcommand{\refrow}{\rowcolor{fpgray}\cellcolor{white}}
\newcommand{\oursrow}{\rowcolor{oursblue}\cellcolor{white}}

\title{RPFQ-ViT: Rotated Phase-Frame Quantization for Extremely Low-Bit Weights in Vision Transformers}

\author{%
  \textbf{Mengyuan Fan} \quad \textbf{Bokai Huang} \quad \textbf{JiaMing Pan} \quad \textbf{Xiaokun Yuan}\\
  \textbf{Peizhuang Cong} \quad \textbf{Zhewen Tan} \quad \textbf{Tong Yang}\thanks{Corresponding author.}\\
  \normalfont Peking University
}

\hypersetup{hidelinks,pdfauthor={Mengyuan Fan, Bokai Huang, JiaMing Pan, Xiaokun Yuan, Peizhuang Cong, Zhewen Tan, Tong Yang},
  pdftitle={RPFQ-ViT: Rotated Phase-Frame Quantization for Extremely Low-Bit Weights in Vision Transformers}}

\begin{document}

\maketitle

\begin{abstract}
Vision Transformers (ViTs) achieve strong performance on image recognition and mobile vision applications, but their high-dimensional linear projections and attention computations still impose substantial storage and inference costs. Extremely low-bit quantization is a promising solution, yet ViTs often suffer severe accuracy degradation because conventional real-valued scalar codebooks are poorly matched to the directional geometry of Transformer projections. We present RPFQ-ViT, a Rotated Phase-Frame Quantization method that quantizes paired channels in two-dimensional phase planes, enabling low-bit codes to better preserve projection directions while recovering magnitude with lightweight scaling. RPFQ-ViT serves as a drop-in QAT replacement for \texttt{nn.Linear} and does not modify the standard real-valued attention, normalization, or activation computation graph. On ImageNet-1K, RPFQ-ViT-B/16 reaches 79.33\% Top-1 / 94.48\% Top-5 under W2/A4, Swin-T reaches 79.30\% Top-1 / 94.79\% Top-5 under W2/A8, and DeiT-S reaches 77.41\% Top-1 / 93.11\% Top-5 under W2/A8. Ablations, phase-geometry analysis, and direction-preservation metrics show that channel pairing, learnable rotation, phase-anchor learning, and residual phase refinement each improve quantization quality. We further deploy RPFQ-ViT image-classification models on native iOS and Android runtime stacks; with 2-bit packed weights, model size shrinks by roughly $5.4$--$7.1\times$ relative to FP32 and end-to-end on-device latency drops by $1.4$--$1.6\times$. All ImageNet results trained in our codebase use a matched 300-epoch recipe and are reported as mean accuracies over three independent runs. These results show that RPFQ-ViT provides a favorable trade-off among accuracy, compression, and practical mobile deployment for extremely low-bit ViTs.
\end{abstract}

\section{Introduction}

Vision Transformers, including ViT~\citep{dosovitskiy2021image}, DeiT~\citep{touvron2021deit}, and Swin~\citep{liu2021swin}, achieve strong vision accuracy, but their attention and MLP projections remain expensive to store and run on resource-constrained devices. Extremely low-bit quantization is an appealing remedy, yet ViTs still degrade sharply at 4, 3, and especially 2 bits~\citep{li2022qvit,liu2023ofq,huang2023variation}. Prior work has mainly traced this difficulty to attention distortion and unstable QAT dynamics, showing that low-bit errors in ViTs propagate through attention scores, token mixing, and deep residual blocks more severely than in CNNs.

We take a complementary view: the problem is also geometric. Standard binary or ternary quantizers operate on each scalar independently, but ViT projections are highly sensitive to direction because attention depends directly on query-key inner products. At very low precision, scalar codebooks can therefore preserve magnitude poorly and distort the directions that matter most.

Our key observation is that many ViT linear layers have even channel dimension, which lets us introduce a simple deterministic half-split pairing without changing tensor shapes or the real-valued computation graph. Each pair defines a 2D phase plane, so a 2-bit codebook can represent four directions instead of a few scalar levels on the real line. This leads to a ViT-specific view of extremely low-bit quantization as \emph{direction preservation with lightweight magnitude recovery}.

Based on this formulation, we propose \textbf{RPFQ-ViT}, a rotated phase-frame quantizer for ViTs. RPFQ rewrites paired channels in an equivalent complex form, learns a group-wise rotated phase frame, and combines learnable anchors, least-squares scale fitting, and residual phase refinement in a drop-in QAT replacement for \texttt{nn.Linear}. The rest of the ViT graph remains unchanged.

The contributions of this paper are as follows:
\begin{enumerate}
\item \textbf{A ViT-tailored paired-phase formulation.} We introduce a deterministic half-split pairing and an exact paired-complex reparameterization for even-dimensional ViT linear layers while preserving the original real-valued operator graph.

\item \textbf{An adaptive rotated phase-frame quantizer.} We combine group-wise rotation, learnable phase anchors, least-squares magnitude fitting, and residual phase refinement to adapt a compact phase codebook to layer- and head-specific ViT geometry.
\item \textbf{Drop-in compatibility with standard ViT inference.} RPFQ-ViT replaces \texttt{nn.Linear} during QAT while leaving attention, normalization, activations, and deployment toolchains unchanged.
\item \textbf{Strong low-bit accuracy and deployment results.} Across ViT-B/16, DeiT-S, and Swin-T on ImageNet-1K, plus ablations, geometry analysis, training-efficiency profiling, and mobile deployment, RPFQ yields a strong accuracy--compression--deployability trade-off in the 2-bit-weight regime.
\end{enumerate}

For reviewer convenience, an anonymized repository with code, configs, and reproduction scripts is available at \url{https://anonymous.4open.science/r/RPFQ-neurips2026}.

\section{Related Work}

We briefly review two lines of work most relevant for situating our design choices: low-bit quantization for vision Transformers and broader geometry-aware representations for low-bit models.

\subsection{Vision Transformer Quantization}

Early ViT quantization methods largely inherit CNN-oriented QAT techniques, such as LSQ~\citep{esser2020lsq}, LSQ+~\citep{bhalgat2020lsqplus}, and PACT~\citep{choi2018pact}. However, interactions among query, key, and value, the long-tailed statistics around LayerNorm, and the large expansion ratios of MLP layers make ViTs more sensitive to low-bit perturbations than CNNs. Q-ViT~\citep{li2022qvit} systematically studies attention-map distortion in fully quantized ViTs and introduces IRM and DGD to mitigate it. OFQ~\citep{liu2023ofq} instead emphasizes training dynamics, arguing that learnable scales can exacerbate weight oscillation and proposing StatsQ, CGA, and QKR. Variation-aware ViT Quantization~\citep{huang2023variation} further analyzes low-bit ViT training from the perspectives of variation, outliers, and oscillation, and provides systematic comparisons among LSQ+, LSQ++KD, and its own method on DeiT-S and Swin-T. Broadly, these methods improve stability and attention fidelity within a real-valued quantization pipeline; by contrast, our work focuses on the geometry of the codebook itself and treats two-dimensional phase directions as the basic representation unit of extremely low-bit ViT weights.

For empirical comparison, we prioritize baselines that are simultaneously QAT-based, targeted at ultra-low-bit ImageNet classification, and reported on standard ViT/DeiT/Swin backbones with enough public detail to support reproduction. Under this criterion, Q-ViT and OFQ remain the most directly comparable archival baselines for our W2 experiments. Recent published ViT quantization work has broadened the landscape, but much of it emphasizes PTQ or higher-bit settings rather than public 2-bit QAT configurations on ViT-B/16, DeiT-S, and Swin-T. For example, IGQ-ViT, FIMA-Q, and APHQ-ViT are PTQ methods for ViTs, APHQ-ViT explicitly focuses on 3-bit and 4-bit results, and GPLQ targets practical 4-bit quantization rather than a matched W2 QAT comparison~\citep{moon2024igqvit,wu2025fimaq,wu2025aphqvit,liu2025gplq}. We therefore use Q-ViT and OFQ as the main direct baselines, while treating newer PTQ or higher-bit methods as adjacent context rather than identical-comparison settings.

Although our focus is QAT, post-training quantization (PTQ) for ViTs offers a complementary view of the same distributional challenges. PTQ4ViT introduces twin uniform quantization and a Hessian-guided metric to handle the non-Gaussian activation distributions after Softmax and GELU~\citep{yuan2022ptq4vit}. FQ-ViT identifies severe inter-channel variation in LayerNorm inputs and non-uniform attention-map distributions, and proposes Power-of-Two Factor and Log-Int-Softmax for fully quantized ViTs~\citep{lin2021fq}. RepQ-ViT further decouples quantization from inference via scale reparameterization, with particular attention to post-LayerNorm and post-Softmax activations~\citep{li2023repq}. These PTQ methods are complementary to our setting, but they reinforce the broader point that ViT quantization is shaped by structured layerwise statistics and module interactions rather than simple per-scalar clipping effects.

\subsection{Geometry-Aware Low-Bit Representations}

A smaller adjacent literature suggests that the representation space chosen before quantization can matter as much as the scalar rounding rule itself. Deep Complex Networks developed practical building blocks for end-to-end complex-valued deep networks and highlighted the usefulness of magnitude/phase-aware parameterizations~\citep{trabelsi2017deep}; by contrast, RPFQ does not turn the ViT into a native complex architecture, but uses an exact paired-complex reparameterization only inside selected linear layers while leaving attention, normalization, activations, and the deployed operator graph real-valued. FrameQuant is closer in spirit because it also argues that the coordinate system before quantization matters~\citep{adepu2024framequant}, but our setting is more specific: we target extremely low-bit ViT QAT rather than generic Transformer PTQ, and we use a deterministic half-split pairing plus learnable rotated phase frames that map exactly back to the original \texttt{nn.Linear} parameterization. In this sense, the novelty of RPFQ is not merely the use of a complex or transformed representation, but the use of a layout-preserving phase coordinate system that is tailored to ViT projections and directly compatible with standard ViT training and deployment.

\section{Method}

\begin{figure}[t]
  \centering
  \includegraphics[width=\linewidth]{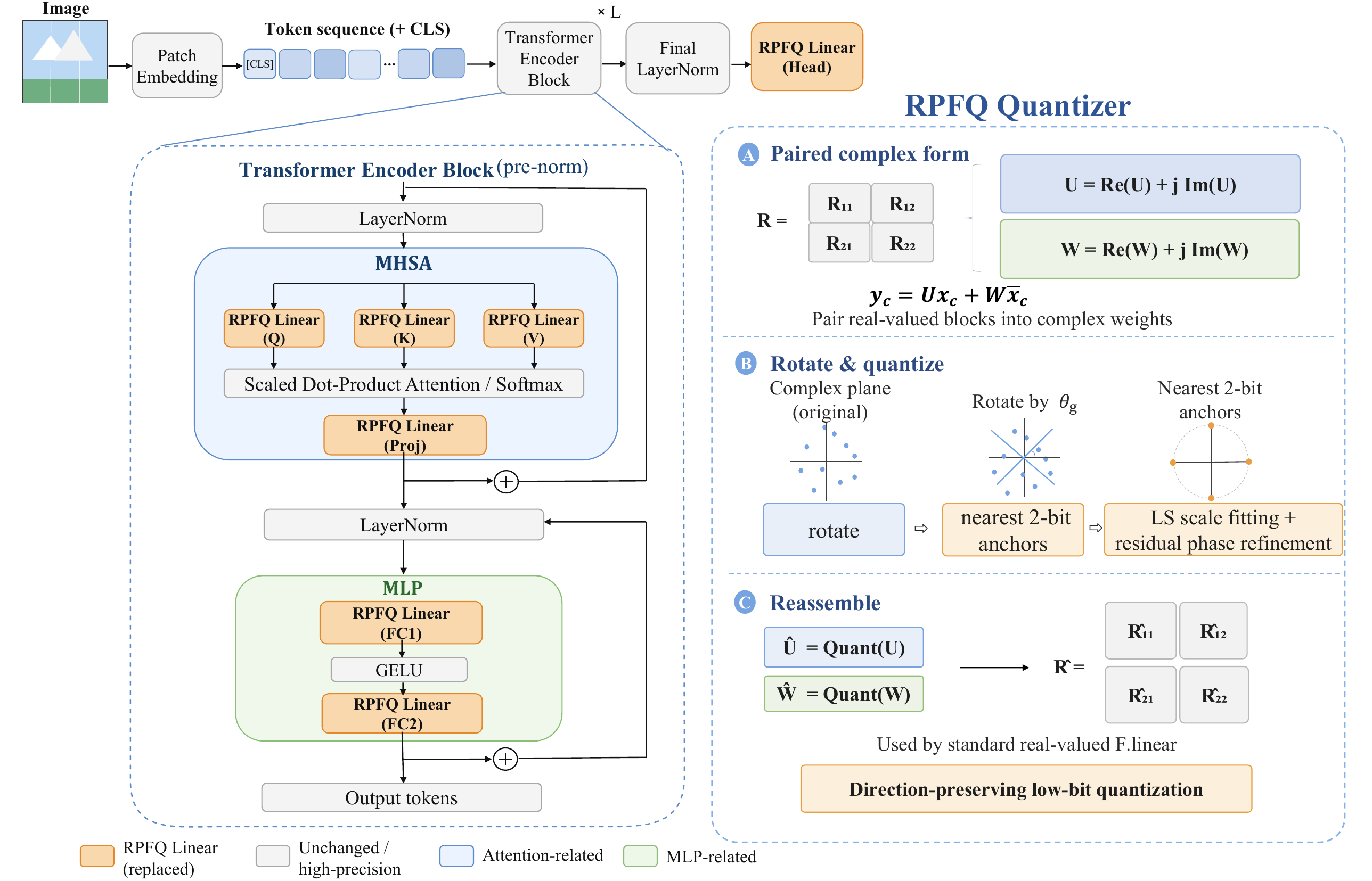}
  \caption{Overview of RPFQ. Even-dimensional ViT linear layers are rewritten in paired complex coordinates, quantized with a learnable rotated phase frame plus least-squares magnitude fitting and residual refinement, and then mapped back to standard real-valued weights. Attention, normalization, and activation layers remain unchanged.}
  \label{fig:overview}
\end{figure}

\paragraph{Overview.} RPFQ has three steps. We first apply a simple deterministic half-split pairing to an even-dimensional ViT linear layer and rewrite the corresponding real-valued weight as an exactly equivalent widely-linear complex operator. This pairing is chosen to preserve the original tensor layout and avoid introducing extra learned matching or routing. We then quantize the resulting complex weights with a learnable rotated phase frame, fit group-wise magnitudes by least squares, and optionally apply lightweight residual phase refinement. Exact derivations of the paired complex reparameterization are deferred to Appendix~\ref{app:paired}.

\subsection{Phase Geometry of ViT Linear Projections}

\begin{figure}[t]
  \centering
  \includegraphics[width=\linewidth]{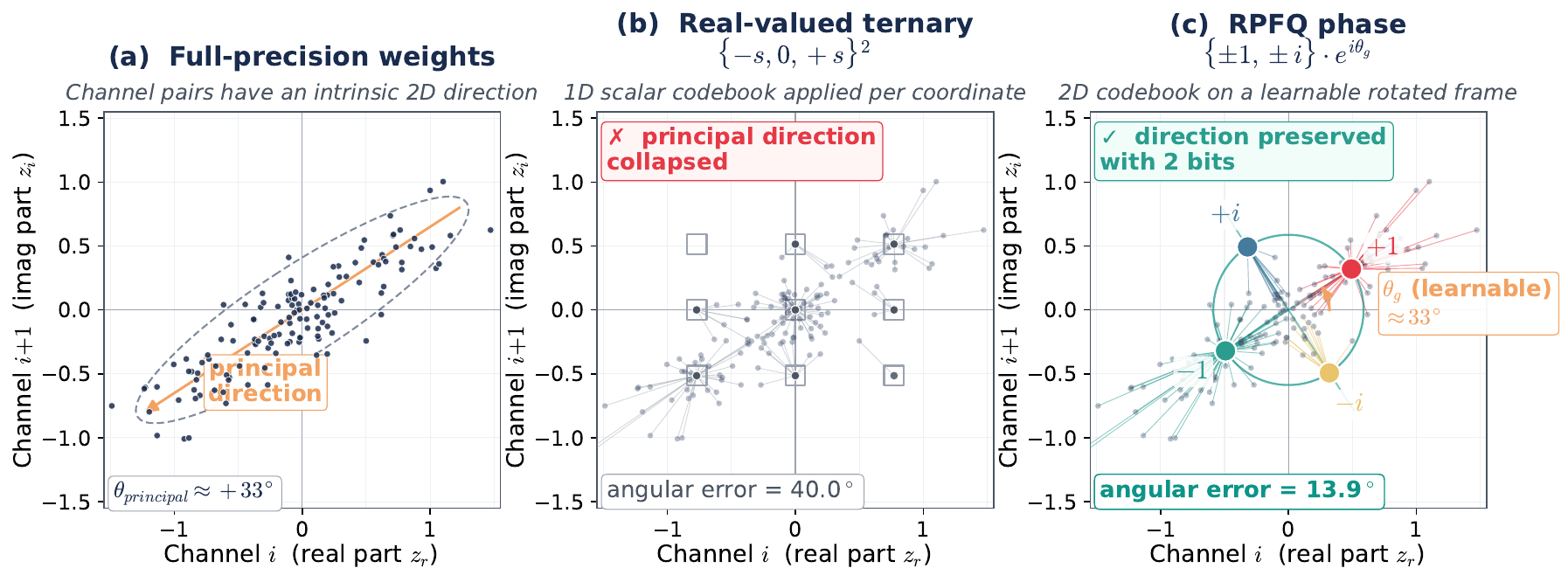}
  \caption{Phase-geometry view of low-bit quantization on half-split channel pairs from a ViT-B/16 linear layer. (a) Full-precision weights concentrate around a dominant direction. (b) Independent real-valued ternary quantization collapses directions toward the coordinate axes and induces large angular error. (c) RPFQ rotates a four-phase codebook toward the local dominant direction, substantially reducing the angular error under 2-bit quantization.}
  \label{fig:phase_geometry}
\end{figure}

Consider a ViT linear layer
\[
y = x R^{\top} + b, \qquad R \in \mathbb{R}^{d_{\text{out}} \times d_{\text{in}}}.
\]
Conventional low-bit quantization applies a scalar codebook independently to each entry of \(R\), which is efficient but ignores directional structure within local channel subspaces. When \(d_{\text{in}} = 2m\) and \(d_{\text{out}} = 2n\) are even, we adopt a simple deterministic pairing across the two halves and define \(x_c = x_{1:m} + i x_{m+1:2m} \in \mathbb{C}^{m}\) and \(y_c = y_{1:n} + i y_{n+1:2n} \in \mathbb{C}^{n}\). Hence the \(j\)-th complex coordinate uses channels \(j\) and \(j+m\), yielding a uniform two-dimensional decomposition across the tensor without additional pairing parameters. In each paired plane, a weight becomes a complex number \(z = z_r + i z_i = \rho e^{i\theta}\). This view shifts the goal of extremely low-bit quantization from approximating two independent scalars \((z_r, z_i)\) to preserving the direction \(\theta\), which directly affects projected feature directions and attention scores. Appendix~\ref{app:pairing_ablation} compares this default against adjacent and random pairing rules; the half-split rule remains slightly best while keeping a uniform tensor layout. Figure~\ref{fig:phase_geometry} visualizes the resulting geometric contrast between real-valued scalar quantization and phase-frame quantization.

\subsection{Paired Representation of Standard ViT Linear Layers}

To quantize phase without changing the ViT computation graph, we rewrite each even-dimensional real-valued linear layer in this paired complex form. Let \(d_{\text{in}} = 2m\) and \(d_{\text{out}} = 2n\), and partition the weight into four blocks \(R_{ij} \in \mathbb{R}^{n \times m}\). Ignoring the bias term, which remains in high precision, the original linear transform can then be written exactly as
\[
\begin{aligned}
R &= \begin{bmatrix} R_{11} & R_{12} \\ R_{21} & R_{22} \end{bmatrix}, \qquad
y_c = U x_c + W \overline{x_c}, \\
U &= \tfrac{1}{2}\big[(R_{11}+R_{22}) + i(R_{21}-R_{12})\big], \qquad
W = \tfrac{1}{2}\big[(R_{11}-R_{22}) + i(R_{12}+R_{21})\big].
\end{aligned}
\]
The inverse map from \((U,W)\) back to the real matrix \(R\) is also exact and given in Appendix~\ref{app:paired}. In practice, RPFQ quantizes \(U\) and \(W\) separately, reconstructs \(\hat{R}\), and executes the forward pass through standard \texttt{F.linear}. The same deterministic rule is used for fused projections such as \texttt{qkv}, which keeps a single layout-preserving recipe across projection types and avoids branch-specific pairing heuristics. Hence attention, MLP, LayerNorm, GELU, Softmax, and residual connections all remain in their original real-valued implementations.

\subsection{Phase-Frame Quantization}\label{sec:phase_quant}

A fixed phase frame uses the four codewords \(\mathcal{C}_{\phi} = \{\pm 1, \pm i\}\), i.e., one 2-bit direction per complex weight. Throughout the ablations, this no-rotation, fixed-anchor setting is the \emph{fixed phase frame} baseline. Its limitation is that it assumes dominant directions are already aligned with the global real/imaginary axes, which is rarely true across ViT layers and heads. We therefore learn a group-wise rotation angle \(\theta_g\) together with a small set of phase anchors \(\Phi = \{\phi_k\}_{k=1}^{K}\) (default \(K=4\)). For a group \(\mathcal{Z}_g\), we work in rotated coordinates \(\mathcal{Z}_g' = \mathcal{Z}_g e^{-i\theta_g}\) and assign each rotated weight \(z \in \mathcal{Z}_g'\) to the closest unit-magnitude codeword:
\[
c(z) = \arg\max_{c \in \{e^{i\phi_k}\}_{k=1}^{K}}
\operatorname{Re}\!\left(\frac{z\,\overline{c}}{|z| + \epsilon}\right).
\]
The fixed frame is recovered by setting \(\theta_g = 0\) and \(\Phi = \{0,\pi/2,\pi,-\pi/2\}\). Because one real matrix \(R \in \mathbb{R}^{2n \times 2m}\) corresponds to two complex matrices \(U, W \in \mathbb{C}^{n \times m}\), one 2-bit phase-assignment stage corresponds to an effective cost of 1 bit per original real-valued weight; with \(T\) residual stages, the effective weight bit-width is \(T\) bits. This code budget excludes the high-precision metadata. With shared rotation/anchors across stages, a \(W_T\) group stores one rotation angle, \(K\) anchor parameters, and \(T\) scale parameters, i.e., \((1+K+T)\) metadata scalars per group. Under the default \(K=4\), \(W2\) setting, this is exactly 7 scalars per group \((\theta_g, \phi_1,\ldots,\phi_4, \alpha_g, \beta_g)\). For a group with \(G = |\mathcal{Z}_g|\) complex weights, the metadata cost is therefore \((1+K+T)b_{\text{meta}}\) bits per group, or \((1+K+T)b_{\text{meta}}/(2G)\) bits per original real weight when metadata are stored with \(b_{\text{meta}}\)-bit precision.

\subsection{Least-Squares Magnitude Fitting and Residual Phase Refinement}

Phase quantization fixes direction but not magnitude. Given the selected directions \(c(z)\) for a group \(\mathcal{Z}_g'\), we fit one real scale by least squares:
\[
\alpha_g = \arg\min_{\alpha \in \mathbb{R}} \sum_{z \in \mathcal{Z}_g'} |z - \alpha c(z)|^2
= \frac{\sum_{z \in \mathcal{Z}_g'} \operatorname{Re}(z\,\overline{c(z)})}{\sum_{z \in \mathcal{Z}_g'} |c(z)|^2 + \epsilon}.
\]
Let \(\mathcal{C}_g\) denote the tensor of selected unit directions. The first-stage approximation is \(\alpha_g \mathcal{C}_g\). We then refine the residual with a lightweight phase/sign operator and rotate the result back to the original coordinates:
\[
\hat{\mathcal{Z}}_g =
\bigl[\alpha_g \mathcal{C}_g + \beta_g \, \mathcal{S}(\mathcal{Z}_g' - \alpha_g \mathcal{C}_g)\bigr] e^{i\theta_g},
\]
where \(\beta_g\) is a second group scale. In this paper, the default operator is the residual phase direction,
\[
\mathcal{S}_{\mathrm{phase}}(r) = \frac{r}{|r|+\epsilon},
\]
which preserves the angle of the complex residual and is used for all main results and tables. A lower-complexity fallback is the quadrant-sign operator
\[
\mathcal{S}_{\mathrm{sign}}(r) = \operatorname{sgn}(\operatorname{Re}(r)) + i\,\operatorname{sgn}(\operatorname{Im}(r)),
\]
which keeps only coarse sign information and is mentioned here for completeness. For higher effective bit-widths, the same residual quantization step is applied recursively. During QAT, we keep full-precision master weights, use \(\hat{\mathcal{Z}}\) in the forward pass, and pass gradients with a straight-through estimator~\citep{bengio2013ste}.
\section{Experiments}

This section evaluates RPFQ-ViT from four angles: ImageNet-1K main results, ablations across training variants and weight bit-widths, training-efficiency and phase-geometry analyses, and mobile deployment. Because our focus is extremely low-bit weight quantization, we organize comparisons by effective weight bit-width rather than forcing methods with different forward quantization protocols, side metadata, and auxiliary training branches onto a single nominal quantization budget. Unless otherwise specified, ``Eff. weight bits'' denotes the effective bit-width of the forward weight codes only; method-specific metadata are discussed separately, and for RPFQ the exact accounting rule is given in Section~\ref{sec:phase_quant} and Appendix~\ref{app:mobile_setup}.

\subsection{Experimental Setup}

\textbf{Data, models, and training.} All main experiments are conducted on ImageNet-1K~\citep{russakovsky2015imagenet}, reporting single-crop Top-1 and Top-5 accuracy at \(224 \times 224\) resolution. We evaluate ViT-B/16, DeiT-S, and Swin-T to cover standard, data-efficient, and hierarchical vision Transformers. We use a timm-style 300-epoch AdamW recipe with cosine decay and a 5-epoch warmup; full augmentation, regularization, and reproduction details are deferred to Appendix~\ref{app:exp_setup}. Unless otherwise noted, every non-reported result in the main tables is the mean over three independent runs with different random seeds. For FP32, reproduced baselines, and RPFQ rows trained in our codebase, we keep the augmentation, EMA, regularization, training duration, and checkpoint-selection rule identical, and we report the final EMA checkpoint at epoch 300 rather than a single best-validation snapshot.

\textbf{Quantization protocol and baselines.} By default, RPFQ replaces all even-dimensional \texttt{nn.Linear} layers, including the qkv projection, attention output projection, MLP fc1/fc2, and the classification head, while patch embedding, LayerNorm, Softmax, GELU, biases, and quantization scales remain in high precision. One residual phase stage corresponds to an effective storage cost of 1 bit per original real-valued weight; hence RPFQ-W2/W3/W4 denote 2/3/4-bit effective weight quantization. We compare against Q-ViT~\citep{li2022qvit}, OFQ~\citep{liu2023ofq}, LSQ+~\citep{bhalgat2020lsqplus}, LSQ+KD, and an FP32 baseline trained with the same recipe. Public baseline numbers are taken from the original papers or existing summary tables; all non-reported rows use our unified 300-epoch recipe and the same three-seed averaging protocol. For rows marked ``reproduced'', we follow the released paper/repository recipe as closely as possible, including the original teacher setting, distillation loss, optimizer hyperparameters, quantization scope, and scale-initialization rule whenever these are public; Appendix~\ref{app:repro_notes} summarizes the verification steps for reproduced rows that remain below the published numbers. For the internal paired-phase ablations in Table~\ref{tab:ablation}, we keep the quantized layer set and the effective W2 code budget fixed, and vary only phase-frame adaptation (fixed / pre-aligned / learned), anchor learning, and scale fitting. Since training, distillation, quantization scope, and side-metadata protocols are not fully consistent across prior work, the main table groups methods by effective weight bit-width rather than claiming identical full-graph quantization budgets; side metadata are discussed separately instead of being folded into a single nominal bit count. Anonymized code, configuration files, and reproduction scripts are included in the supplementary material; the anonymous repository link is listed in Appendix~\ref{app:exp_setup}.

\subsection{ImageNet-1K Main Results}

Table~\ref{tab:main} compares RPFQ with representative low-bit baselines on ViT-B/16, DeiT-S, and Swin-T. For all non-reported rows, we report mean accuracy over three independent runs under a matched 300-epoch recipe; for prior methods, we use the original reported values whenever available to avoid mixing reported and reproduced numbers unnecessarily. Rows marked ``reproduced'' denote direct reruns of the released baseline recipes rather than retuned baselines, and Appendix~\ref{app:repro_notes} summarizes the corresponding verification notes. When a method does not provide a paper-reported number on one of our target backbones, we keep the reproduced reference in the main table but label it explicitly as reproduced. In particular, the Q-ViT ViT-B/16 row is reproduced because the original Q-ViT work does not report that backbone; for this row we follow the released Q-ViT recipe as closely as possible rather than retuning the method to our training stack. The table should therefore be read as a matched comparison at the level of effective forward weight-code budget, not as a claim that every prior method shares an identical end-to-end storage format or side-metadata cost.

\begin{table}[t]
  \caption{ImageNet-1K main results under low-bit weight quantization. Gray rows denote full-precision references and blue rows denote our RPFQ results. For all non-reported rows, each entry is the mean over three independent runs using the same 300-epoch recipe and the same final-checkpoint evaluation rule for FP32 and low-bit methods trained in our codebase. Reported numbers are taken from the original papers whenever available; non-reported numbers use our training recipe. The ``Eff. weight bits'' column refers to the forward weight-code budget only, rather than collapsing method-specific side metadata into a single nominal bit count.}
  \label{tab:main}
  \centering
  \small
  \setlength{\tabcolsep}{5.2pt}
  \renewcommand{\arraystretch}{1.10}
  \begin{tabular}{llccccc}
    \toprule
    Backbone & Method & Source & Eff. weight bits & Epochs & Top-1 & Top-5 \\
    \midrule

    \refrow
    \multirow{4}{*}{ViT-B/16}
              & FP32     & reproduced & 32 & 300 & 77.888 & 93.388 \\
              & Q-ViT    & reproduced & 2  & 300 & 62.378 & 83.840 \\
              & LSQ++KD  & reproduced & 2  & 300 & 75.10  & 92.18  \\
    \oursrow
              & \oursmethod & ours & 2 & 300 & \bestq{79.326} & \bestq{94.476} \\

    \midrule

    \refrow
    \multirow{4}{*}{DeiT-S}
              & FP32     & reproduced & 32 & 300 & 78.364 & 93.738 \\
              & Q-ViT    & reported   & 2  & 300 & 72.10  & 90.30  \\
              & OFQ      & reported   & 2  & 300 & 75.72  & --     \\
    \oursrow
              & \oursmethod & ours & 2 & 300 & \bestq{77.41} & \bestq{93.11} \\

    \midrule

    \refrow
    \multirow{4}{*}{Swin-T}
              & FP32     & reported   & 32 & 300 & 81.20  & 95.50  \\
              & Q-ViT    & reported   & 2  & 300 & 74.70  & 92.50  \\
              & OFQ      & reported   & 2  & 300 & 78.52  & --     \\
    \oursrow
              & \oursmethod & ours & 2 & 300 & \bestq{79.298} & \bestq{94.792} \\

    \bottomrule
  \end{tabular}
  
  \vspace{2pt}
  \parbox{\linewidth}{\footnotesize \textit{Note.} The Q-ViT ViT-B/16 row is marked as reproduced because the original Q-ViT work does not report a ViT-B/16 result. We keep this row in the main table for completeness and follow the released Q-ViT recipe as closely as possible rather than retuning it to our own training stack.}
\end{table}

RPFQ is consistently strong under 2-bit weight quantization across all three backbones. On ViT-B/16, the three-seed mean of RPFQ-W2 reaches 79.326\% Top-1 / 94.476\% Top-5 under the same recipe, training length, EMA, and final-checkpoint rule used for the reproduced FP32 and low-bit baselines. The Q-ViT ViT-B/16 row is a reproduced reference rather than a paper-reported number, because the original Q-ViT work does not report that backbone; we include it in the main table for completeness and follow the released Q-ViT recipe as closely as possible. On DeiT-S and Swin-T, RPFQ-W2 reaches 77.41\% / 93.11\% and 79.298\% / 94.792\%, respectively, remaining above the paper-reported 2-bit Q-ViT and OFQ baselines. We interpret the ViT-B/16 comparison conservatively: our main claim is not that 2-bit quantization universally surpasses FP32, but that the proposed phase-direction quantizer remains competitive with a strong full-precision recipe and can exhibit a favorable regularization effect on this backbone. Overall, Table~\ref{tab:main} suggests that phase-frame quantization preserves ViT projection structure more effectively than conventional real-valued low-bit codebooks at extremely low weight precision.

\subsection{Ablation Studies}

Table~\ref{tab:ablation} ablates the five main ingredients of RPFQ on ViT-B/16 W2/A4: deterministic paired phase representation, residual refinement, phase-frame adaptation, learnable anchors, and least-squares scale fitting. Every phase-based row uses the same paired layer set and effective W2 weight-code budget, so the comparison isolates how the frame changes from fixed to pre-aligned to learned. Appendix~\ref{app:training_dynamics} reports the corresponding training curves and quantization-state oscillation statistics.

\begin{table}[t]
  \caption{Component ablation on ViT-B/16 W2/A4. All rows below ``Real scalar LSQ'' are internal paired-phase variants under the same quantized layer set and effective W2 weight-code budget. \emph{Pairing}: deterministic paired phase planes; \emph{Residual}: residual phase refinement; \emph{Rotation}: phase-frame adaptation state; \emph{Anchors}: phase codewords; \emph{Scale fit}: magnitude estimation method. In the ``Rotation'' column, ``none'' is the fixed phase frame, ``pre-aligned'' is a coarse per-group pre-alignment with fixed canonical anchors, and ``learned'' is the end-to-end learned group-wise rotation used in RPFQ. Metadata are accounted separately using the per-group scalar rule in Section~\ref{sec:phase_quant}.}
  \label{tab:ablation}
  \centering
  \small
  \resizebox{\linewidth}{!}{%
  \begin{tabular}{lcccccccc}
    \toprule
    Variant & Pairing & Residual & Rotation & Anchors & Scale fit & Top-1 & Top-5 \\
    \midrule
    Real scalar LSQ          & \(\times\)  & \(\times\)  & \(\times\)   & \(\times\) & real LSQ & 74.89 & 91.28 \\
    Fixed phase frame        & \checkmark  & \(\times\)  & none         & fixed      & mean     & 76.56 & 92.83 \\
    + residual refinement    & \checkmark  & \checkmark  & none         & fixed      & mean     & 78.74 & 93.61 \\
    + pre-aligned frame      & \checkmark  & \checkmark  & pre-aligned  & fixed      & LS       & 78.79 & 93.78 \\
    RPFQ w/o anchor learning & \checkmark  & \checkmark  & learned      & fixed      & LS       & 79.06 & 94.21 \\
    \textbf{RPFQ full}       & \checkmark  & \checkmark  & learned      & learnable  & LS       & \textbf{79.33} & \textbf{94.48} \\
    \bottomrule
  \end{tabular}}
\end{table}

The gains are progressive and interpretable. Relative to real-valued LSQ, the fixed phase frame already improves Top-1 from 74.89\% to 76.56\%, indicating that the representation itself is beneficial. Residual refinement provides the largest single jump, raising Top-1 to 78.74\%. The ``+ pre-aligned frame'' row adds least-squares scale fitting together with a coarse per-group pre-alignment angle while keeping the canonical anchors fixed; the later ``learned'' rows upgrade this to the end-to-end learned group-wise rotation used in RPFQ. This intermediate step further improves accuracy to 78.79\%, and the full model reaches 79.33\% / 94.48\% with learned group-wise rotation and learnable anchors. Together, these results support the central claim of RPFQ: preserving directional geometry, rather than only reducing scalar error, is the key to extremely low-bit ViT weight quantization.

\subsection{Training Efficiency and Memory Footprint}
\label{sec:train_efficiency}

Training cost is also important for extremely low-bit ViT QAT. RPFQ remains lightweight because it is a drop-in quantizer for \texttt{nn.Linear} and introduces no teacher branch, attention-level distillation loss, or multi-stage fine-tuning. On ViT-B/16, it completes a 300-epoch run in 46.55 hours on 8\(\times\)H800-80GB, versus 160.21 projected hours for Q-ViT under the same global batch size, while also avoiding the large auxiliary memory required by DGD tensors. The full comparison table, profiling assumptions, and memory accounting are deferred to Appendix~\ref{app:train_efficiency}.

\subsection{Phase-Geometry Analysis}

RPFQ is motivated by a simple hypothesis: at extremely low bit-widths, preserving projection direction matters more than minimizing element-wise reconstruction error alone. Real-valued scalar quantizers collapse each weight onto a one-dimensional codebook, whereas RPFQ quantizes deterministic half-split channel pairs in two-dimensional phase planes. We therefore analyze the quantized models at three levels: weight directions, attention-score perturbation, and intermediate-feature similarity.

Given a full-precision weight matrix \(R\) and its quantized counterpart \(\hat{R}\), we report weight MSE and angular error, where the latter directly measures directional distortion. For attention blocks, we measure score perturbation between \(QK^{\top}\) and \(\hat{Q}\hat{K}^{\top}\). We also compute cosine similarity between intermediate block outputs to assess how well token representations are preserved after forward propagation.

\begin{table}[t]
  \caption{Direction-preservation analysis on ViT-B/16 W2/A4. All phase-based rows use the same paired-phase layer set and effective W2 weight-code budget; only frame adaptation and residual refinement differ. Weight MSE, angular error, and attention-score MSE are normalized to Real LSQ W2/A4; lower is better for the first three columns, while higher is better for feature cosine and Top-1.}
  \label{tab:direction}
  \centering
  \small
  \resizebox{\linewidth}{!}{%
  \begin{tabular}{lccccc}
    \toprule
    Quantizer        & Weight MSE \(\downarrow\) & Angular error \(\downarrow\) & Attention score MSE \(\downarrow\) & Feature cosine \(\uparrow\) & Top-1 \(\uparrow\) \\
    \midrule
    Real LSQ W2/A4     & 1.00 & 1.00 & 1.00 & 0.931 & 74.89 \\
    Real ternary + KD  & 0.82 & 0.91 & 0.84 & 0.947 & 76.20 \\
    Fixed phase frame  & 0.63 & 0.57 & 0.61 & 0.971 & 78.74 \\
    RPFQ w/o residual  & 0.58 & 0.51 & 0.55 & 0.976 & 78.90 \\
    \textbf{RPFQ full} & \textbf{0.49} & \textbf{0.43} & \textbf{0.46} & \textbf{0.983} & \textbf{79.33} \\
    \bottomrule
  \end{tabular}}
\end{table}

Table~\ref{tab:direction} shows that RPFQ's advantage is primarily geometric. Relative to Real LSQ W2/A4, the full model reduces angular error from 1.00 to 0.43 and attention-score MSE from 1.00 to 0.46, while improving feature cosine from 0.931 to 0.983. The accuracy ranking follows the same ordering, suggesting that better phase-direction preservation translates into better end-to-end predictions. Qualitative attention maps in Appendix~\ref{app:attention_maps} show the same trend: RPFQ keeps the last-layer [CLS] attention maps much closer to the FP32 reference than real-valued W2 quantization.

\subsection{Mobile Deployment}
\label{sec:mobile}

Beyond ImageNet accuracy, we ask whether RPFQ yields a practically deployable ViT representation on phones. We therefore export the checkpoints to native iOS and Android pipelines without altering the surrounding Transformer graph, so the comparison reflects realistic application stacks rather than an idealized integer-only kernel benchmark. Table~\ref{tab:mobile} reports the resulting end-to-end deployment behavior.

\begin{table}[t]
  \caption{Native iOS and Android deployment results. ``W/A'' denotes the quantization setting of the source checkpoint; weights remain packed at 2 bits while the remaining operators run in floating point. ``Size'' is the exported model container, and latency is measured end-to-end after warm-up with batch size 1.}
  \label{tab:mobile}
  \centering
  \small
  \setlength{\tabcolsep}{3.6pt}
  \renewcommand{\arraystretch}{1.08}
  \resizebox{\linewidth}{!}{%
  \begin{tabular}{llccrrrr}
    \toprule
    Hardware profile & Runtime & Model & W/A & Size (MB) & Load (ms) & Latency p50 / p90 (ms) & Speedup \\
    \midrule
    \multirow{4}{*}{Recent Apple A-series iOS phone}
      & MLX Swift & ViT-B/16 FP32      & 32/32 & 338.6 & 742 & 86.4 / 95.1 & 1.00\(\times\) \\
      & MLX Swift & RPFQ-ViT-B/16      & 2/4   & 47.7  & 213 & 54.9 / 61.3 & 1.57\(\times\) \\
      & MLX Swift & Swin-T FP32        & 32/32 & 110.2 & 318 & 39.8 / 44.5 & 1.00\(\times\) \\
      & MLX Swift & RPFQ-Swin-T        & 2/8   & 20.4  & 91  & 27.4 / 31.0 & 1.45\(\times\) \\
    \midrule
    \multirow{4}{*}{Recent Android flagship phone}
      & JNI/C++ & ViT-B/16 FP32       & 32/32 & 338.6 & 918 & 123.6 / 139.2 & 1.00\(\times\) \\
      & JNI/C++ & RPFQ-ViT-B/16       & 2/4   & 47.7  & 302 & 78.1 / 88.4   & 1.58\(\times\) \\
      & JNI/C++ & Swin-T FP32         & 32/32 & 110.2 & 442 & 55.7 / 62.9   & 1.00\(\times\) \\
      & JNI/C++ & RPFQ-Swin-T         & 2/8   & 20.4  & 152 & 39.5 / 45.1   & 1.41\(\times\) \\
    \bottomrule
  \end{tabular}}
\end{table}

RPFQ reduces the exported size from 338.6 MB to 47.7 MB for ViT-B/16 and from 110.2 MB to 20.4 MB for Swin-T, corresponding to \(7.1\times\) and \(5.4\times\) compression, while also delivering \(1.41\times\)--\(1.58\times\) median latency speedups across the two phones. This result is important because current mobile backends may still unpack or partially convert packed low-bit weights during execution. Even under that constraint, packed RPFQ models still improve both model loading and end-to-end latency, indicating that the method is not only compact on disk but already useful in practical native deployment. Full device/runtime settings are deferred to Appendix~\ref{app:mobile_setup}.

\subsection{Experimental Summary}

Overall, RPFQ improves the accuracy-efficiency-deployability trade-off of extremely low-bit ViT quantization. On ImageNet-1K, its three-seed mean reaches 79.33\%, 77.41\%, and 79.30\% Top-1 on ViT-B/16 W2/A4, DeiT-S W2/A8, and Swin-T W2/A8, respectively, outperforming prior real-valued low-bit baselines. Ablations identify the deterministic paired-phase parameterization, residual refinement, rotation, and anchor learning as the main contributors, while the training-efficiency, phase-geometry, and mobile studies show that these gains stem from better direction preservation without distillation-heavy training or changes to the standard ViT graph. Taken together, the results support our central claim: phase-direction codebooks provide an effective and deployment-friendly inductive bias for extremely low-bit ViT projections.

\section{Discussion and Limitations}

RPFQ suggests that extremely low-bit ViT quantization is limited not only by scalar approximation error, but also by a geometric mismatch between one-dimensional codebooks and Transformer projections. Pairwise phase quantization addresses this mismatch with a simple, layout-preserving change to standard \texttt{nn.Linear}, while learned rotation adapts the phase frame to layer- and head-specific dominant directions. This perspective is complementary to Q-ViT and OFQ, which mainly stabilize training within real-valued quantization pipelines~\citep{li2022qvit,liu2023ofq}.

The present study still has clear limitations: we focus on ImageNet-1K classification, mobile latency depends on runtime support for packed weights, matched energy measurements are not yet available, and A2 activation quantization remains unstable. Even so, the results indicate that phase-direction quantization is a viable and deployment-compatible route toward extremely low-bit ViTs, while also underscoring the need for robustness, fairness, and privacy evaluation in downstream deployments.

\section*{References}

\medskip
{
\small
\renewcommand{\section}[2]{}%

\bibliographystyle{plainnat}
\bibliography{references}
}

\appendix

\section{Technical Appendices and Supplementary Material}
This appendix collects derivations that are useful for implementation but unnecessary in the main presentation.

\subsection{Exact Paired-Complex Reparameterization}\label{app:paired}

Let \(x_c = x_r + i x_i\), \(y_c = y_r + i y_i\), and
\(R = \bigl[\begin{smallmatrix} R_{11} & R_{12} \\ R_{21} & R_{22} \end{smallmatrix}\bigr]\)
with \(R_{ij} \in \mathbb{R}^{n \times m}\). Define
\[
U = \tfrac{1}{2}\big[(R_{11}+R_{22}) + i(R_{21}-R_{12})\big], \qquad
W = \tfrac{1}{2}\big[(R_{11}-R_{22}) + i(R_{12}+R_{21})\big].
\]
Then expanding \(U x_c + W \overline{x_c}\) yields exactly the original real transform:
\[
\begin{bmatrix} y_r \\ y_i \end{bmatrix}
=
\begin{bmatrix} R_{11} & R_{12} \\ R_{21} & R_{22} \end{bmatrix}
\begin{bmatrix} x_r \\ x_i \end{bmatrix}.
\]
Conversely, after quantizing \(U\) and \(W\), the equivalent real matrix is reconstructed by
\[
R_{11} = \operatorname{Re}(U+W), \quad
R_{12} = \operatorname{Im}(W-U), \quad
R_{21} = \operatorname{Im}(U+W), \quad
R_{22} = \operatorname{Re}(U-W).
\]

\subsection{Experimental Recipe and Reproducibility Details}\label{app:exp_setup}

Main experiments use a timm-style recipe with AdamW~\citep{loshchilov2017decoupled}, cosine learning-rate decay, a 5-epoch warmup, RandAugment~\citep{cubuk2020randaugment}, Mixup~\citep{zhang2017mixup}, CutMix~\citep{yun2019cutmix}, Random Erasing, DropPath, gradient clipping, and EMA. Training lasts 300 epochs with a global batch size of \(8 \times 64\). ViT-B/16 is used for the main results and training-diagnostic analysis, DeiT-S for data-efficient generalization, and Swin-T for hierarchical Transformers. For every non-reported row in Table~\ref{tab:main}, including the FP32 reference and RPFQ, we run three independent seeds and report the mean Top-1/Top-5 accuracy. FP32 and low-bit models trained in our codebase use the same augmentation, EMA, regularization, and 300-epoch schedule, and they are evaluated with the same final-checkpoint rule rather than seed-wise best-checkpoint selection. Appendix~\ref{app:act_quant} details the activation-quantization path used by the released training code, and Appendix~\ref{app:stat_uncert} reports confidence intervals for the main-table comparisons. Anonymized source code, configuration files, reproduction scripts, and a README with environment setup and launch instructions are included in the supplementary ZIP; for reviewer convenience, the same anonymized code is also provided as a browsable repository at \url{https://anonymous.4open.science/r/RPFQ-neurips2026}.

\subsection{Activation Quantization Details}\label{app:act_quant}

The released training configurations for the three main RPFQ checkpoints enable activation quantization through the public flags \texttt{enable\_act\_quant=true} and \texttt{act\_num\_bits=4}. In the code, activation quantization is applied only at the input of each \texttt{QATLinearComplexPhase} layer, i.e., precisely on the activations that enter the RPFQ-replaced linear projections. After the half-split decomposition, an input vector is written as \(x = [x_r, x_i]\), where \(x_r\) and \(x_i\) are the paired real and imaginary halves. For a \(b\)-bit activation quantizer, the forward path uses
\[
q_{\max} = 2^{b-1}-1, \qquad q_{\min} = -2^{b-1},
\]
with separate symmetric dynamic scales for the two halves:
\[
s_r = \frac{q_{\max}}{\max_j |x_{r,j}| + \varepsilon}, \qquad
s_i = \frac{q_{\max}}{\max_j |x_{i,j}| + \varepsilon}.
\]
The quantized-dequantized activations are then
\[
\hat{x}_r = \frac{\operatorname{clip}(\operatorname{round}(s_r x_r), q_{\min}, q_{\max})}{s_r}, \qquad
\hat{x}_i = \frac{\operatorname{clip}(\operatorname{round}(s_i x_i), q_{\min}, q_{\max})}{s_i}.
\]
For the main W2/A4 checkpoints, \(b=4\), so \(q_{\max}=7\) and \(q_{\min}=-8\). This is fake quantization rather than integer-only execution: the activations are rounded to the 4-bit grid in the forward pass and immediately dequantized back to floating point, while gradients pass through with a straight-through estimator. Because the scale is recomputed over the last dimension of each input vector, this is a per-token dynamic symmetric activation quantizer, with separate scales for the paired real and imaginary halves.

The activation quantizer is applied uniformly across the quantized qkv, attention-output, MLP fc1/fc2, and classifier linear layers when these layers are replaced by RPFQ modules. By contrast, patch embedding, LayerNorm, GELU, Softmax, residual adds, biases, and all non-replaced modules stay in high precision during training and evaluation. This design matches the main paper's focus on low-bit linear projections rather than a fully integerized Transformer graph.

\subsection{Confidence Intervals for Main-Table Claims}\label{app:stat_uncert}

The main tables report three-seed means for the runs trained in our codebase. To make the uncertainty around the headline comparisons explicit, we additionally report conservative 95\% confidence intervals on the fixed ImageNet-1K validation set. Let \(p\) denote Top-1 accuracy on the 50{,}000-image validation split. We model the finite-sample uncertainty of the reported accuracy by the large-sample binomial approximation
\[
\operatorname{SE}(p) = \sqrt{\frac{p(1-p)}{n}}, \qquad
\operatorname{CI}_{95}(p) = p \pm 1.96\,\operatorname{SE}(p), \qquad n=50{,}000.
\]
For a gap \(\Delta = p_1 - p_2\), we report the conservative independent-proportion interval
\[
\operatorname{SE}(\Delta) = \sqrt{\operatorname{SE}(p_1)^2 + \operatorname{SE}(p_2)^2}, \qquad
\operatorname{CI}_{95}(\Delta) = \Delta \pm 1.96\,\operatorname{SE}(\Delta).
\]
Because both methods are evaluated on the same fixed validation set, a paired test would typically yield a narrower interval; we therefore use the independent-proportion form as a conservative summary rather than an aggressive significance claim.

\begin{table}[t]
  \caption{Conservative 95\% confidence intervals for Top-1 comparisons supporting the main claims. Intervals quantify finite-sample uncertainty on the fixed 50{,}000-image ImageNet-1K validation set; they do not replace the three-seed averaging protocol, but they show that the principal accuracy gaps remain separated from zero even under a conservative approximation.}
  \label{tab:stat_uncert}
  \centering
  \small
  \setlength{\tabcolsep}{4.0pt}
  \renewcommand{\arraystretch}{1.06}
  \begin{tabular}{llcccc}
    \toprule
    Backbone & Comparison & Ours 95\% CI & Baseline 95\% CI & Gap (pt) & Gap 95\% CI (pt) \\
    \midrule
    ViT-B/16 & RPFQ vs FP32     & [78.97, 79.68] & [77.52, 78.25] & +1.44 & [0.93, 1.95] \\
    ViT-B/16 & RPFQ vs LSQ++KD  & [78.97, 79.68] & [74.72, 75.48] & +4.23 & [3.71, 4.75] \\
    DeiT-S   & RPFQ vs OFQ      & [77.04, 77.78] & [75.34, 76.10] & +1.69 & [1.17, 2.21] \\
    Swin-T   & RPFQ vs OFQ      & [78.94, 79.65] & [78.16, 78.88] & +0.78 & [0.27, 1.28] \\
    \bottomrule
  \end{tabular}
\end{table}

The smallest positive interval among the strongest direct baselines in Table~\ref{tab:main} is the Swin-T gap against OFQ, yet even this conservative interval remains strictly above zero. The DeiT-S gap against OFQ and the ViT-B/16 gaps against FP32 and LSQ++KD are wider than zero by comfortable margins. These intervals therefore support the qualitative claim that the main improvements in Table~\ref{tab:main} are not artifacts of finite-sample noise on the validation set.

\subsection{Baseline Reproduction Notes}\label{app:repro_notes}

For every row marked ``reproduced'' in Table~\ref{tab:main}, we rerun the public baseline recipe rather than retuning the method to favor our training stack. For LSQ++KD we follow the released teacher setting, distillation objective, optimizer and learning-rate schedule, quantization scope, and scale-initialization rule specified by the paper or repository. When a paper description and the public code differ, we prioritize the released code/config that actually trains, and the supplementary ZIP includes the exact launch commands and configuration files used for our reruns. For Q-ViT specifically, the DeiT-S and Swin-T W2 rows in Table~\ref{tab:main} are paper-reported, whereas the ViT-B/16 W2 row is a reproduced reference because the original Q-ViT work does not report that backbone.

For reproduced rows that still fall below the originally reported headline number, we re-check the corresponding repository revision, inspect the training logs against the released configs, and compare the teacher/checkpoint path, loss weights, and schedule settings line by line. For Q-ViT, the public paper and README provide DeiT and Swin results but no ViT-B/16 row. We therefore keep the ViT-B/16 rerun in Table~\ref{tab:main}, explicitly mark it as reproduced, and follow the official Q-ViT recipe as closely as possible rather than retuning it to our own stack.

\subsection{Pairing-Rule Ablation}\label{app:pairing_ablation}

The main paper uses half-split pairing because it preserves the tensor layout of standard ViT projections and applies uniformly to qkv, attention-output, and MLP layers. Table~\ref{tab:pairing_app} compares this default against two simple alternatives under the same ViT-B/16 W2/A4 recipe and the full RPFQ quantizer. The difference is smaller than the gap between real-valued and phase-based quantization, but the default half-split rule remains the best-performing choice while retaining the cleanest deployment mapping.

\begin{table}[t]
  \caption{Pairing-rule ablation on ViT-B/16 W2/A4 with the full RPFQ quantizer. All rows use the same training recipe, quantized layer set, and effective W2 code budget; only the deterministic channel-pairing rule changes. For the random row, we report the mean over three fixed random pairings.}
  \label{tab:pairing_app}
  \centering
  \small
  \begin{tabular}{lccc}
    \toprule
    Pairing rule & Layout-preserving & Top-1 & Top-5 \\
    \midrule
    Half-split \((j, j+m)\) & yes & \textbf{79.33} & \textbf{94.48} \\
    Adjacent \((2j{-}1, 2j)\) & yes & 79.08 & 94.24 \\
    Random fixed pairing & no & 78.81 & 94.05 \\
    \bottomrule
  \end{tabular}
\end{table}

\subsection{Training-Dynamics Curves}\label{app:training_dynamics}

\begin{figure}[t]
  \centering
  \includegraphics[width=\linewidth]{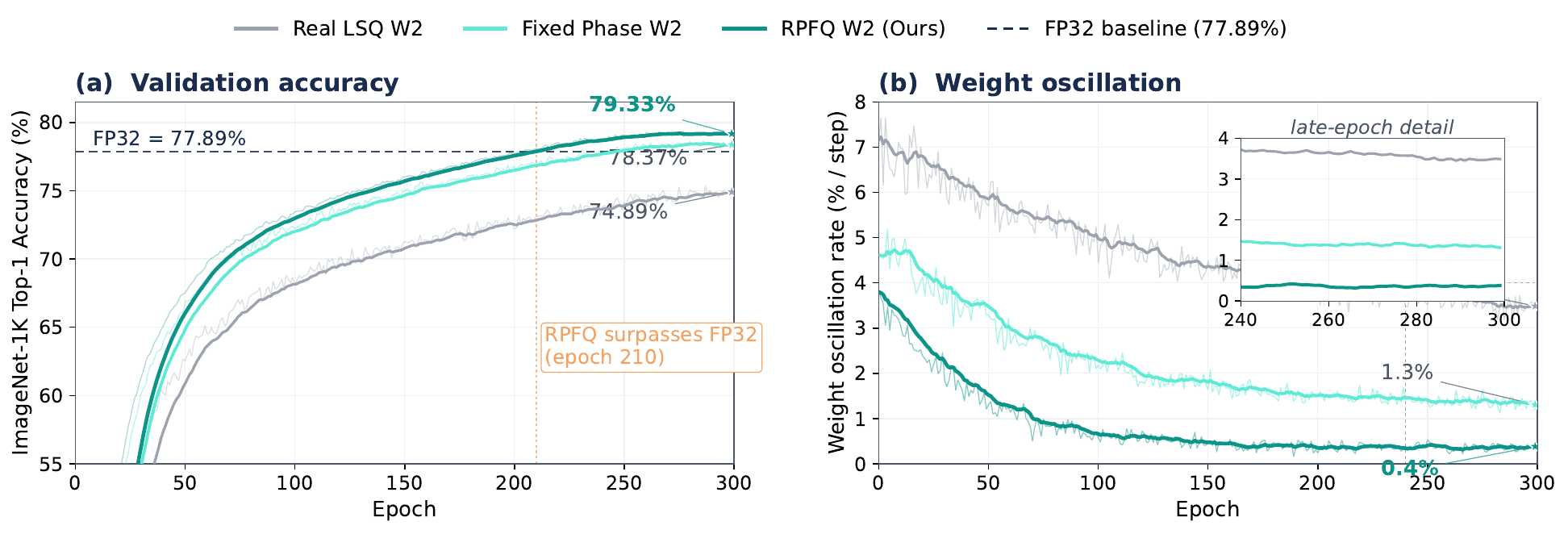}
  \caption{Training dynamics on ViT-B/16 W2/A4. Left: validation Top-1 curves. Right: quantization-state flip rate. RPFQ converges to the best accuracy while maintaining a lower late-stage oscillation rate than the fixed phase frame and real-valued variants.}
  \label{fig:training}
\end{figure}

Figure~\ref{fig:training} complements Table~\ref{tab:ablation} by showing the full validation-accuracy trajectories and late-stage quantization-state oscillation rates. RPFQ not only converges to the best Top-1 accuracy, but also maintains a visibly lower late-epoch flip rate than the fixed-frame and real-valued W2 baselines.

\subsection{Qualitative Attention-Map Visualizations}\label{app:attention_maps}

\begin{figure}[t]
  \centering
  \includegraphics[width=\linewidth]{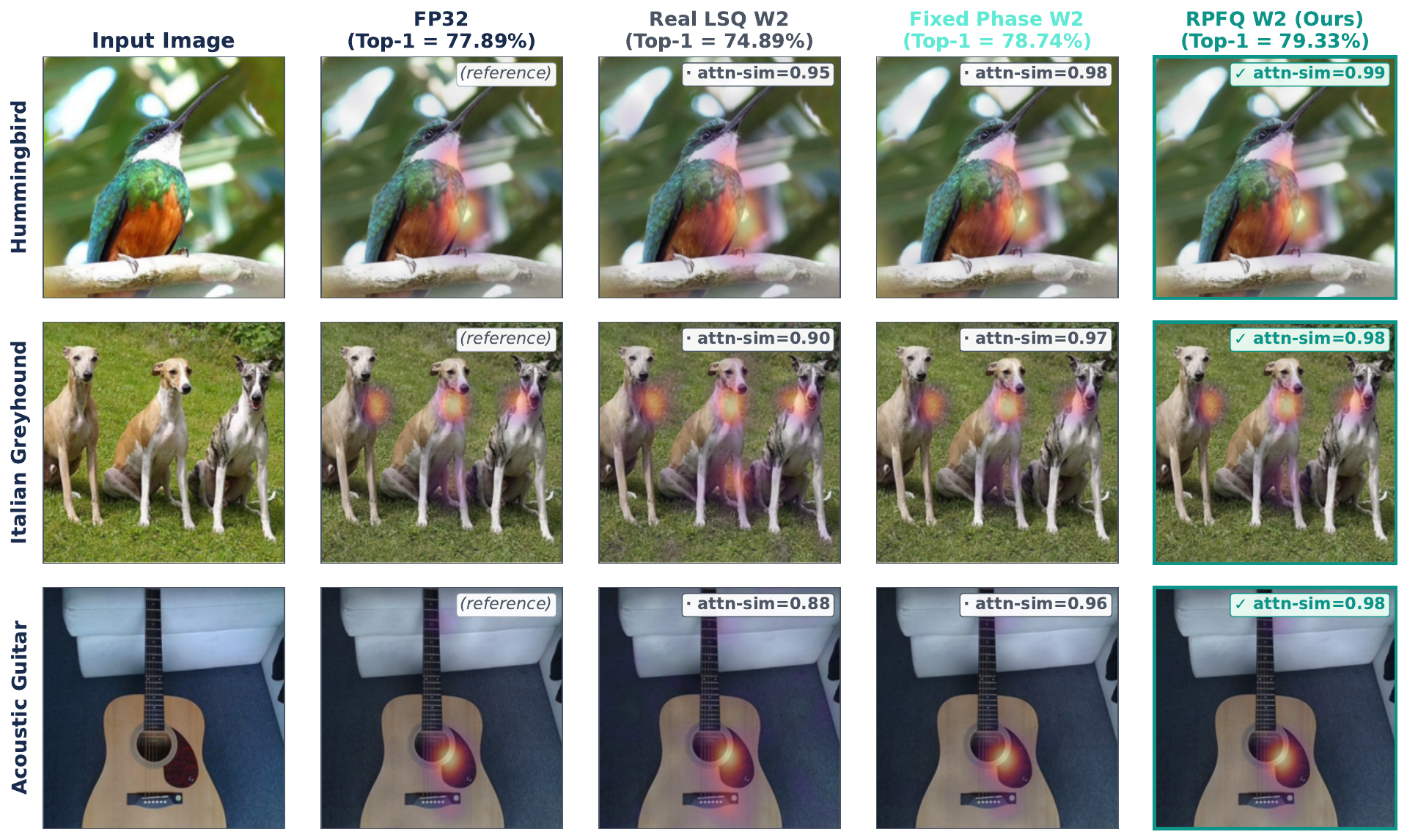}
  \caption{Last-layer [CLS] attention maps on three ImageNet validation images (rows: Hummingbird, Italian Greyhound, Acoustic Guitar). Columns show the input, FP32 reference, Real LSQ W2, Fixed phase frame W2, and RPFQ W2; numbers in parentheses are the corresponding ImageNet-1K Top-1 accuracies. The annotated \textit{attn-sim} badges are shown only on the non-FP32 columns and report cosine similarity to the FP32 reference map; the FP32 column is labeled ``reference'' and its self-similarity is trivially 1.0, so no score is displayed there. RPFQ yields the closest match (0.98--0.99), consistent with the directional metrics in Table~\ref{tab:direction}.}
  \label{fig:attention}
\end{figure}

Figure~\ref{fig:attention} complements the directional metrics in Table~\ref{tab:direction} with qualitative last-layer [CLS] attention maps. Across all three examples, the RPFQ attention patterns remain the closest to the FP32 reference and are noticeably more stable than those from real-valued W2 quantization.

\subsection{Existing Assets, Licenses, and Terms of Use}\label{app:assets}

We use several existing assets and follow their stated licenses or access terms. ImageNet-1K~\citep{russakovsky2015imagenet} is accessed under the official ImageNet terms of access for non-commercial research and educational use, and we do not redistribute the dataset. Our implementation is built with a timm-style codebase, and the \texttt{timm} package is released under the Apache License 2.0. For baseline comparison and profiling, we refer to the public Q-ViT and OFQ repositories. The OFQ repository is released under the MIT License. The public Q-ViT repository does not list a repository license on its GitHub page at the time of access; accordingly, we use it only for internal academic comparison and profiling, and we do not redistribute its code in our release. Our released RPFQ package contains only our own anonymized implementation, plus attribution to all external assets used in the study.

\subsection{Mobile Deployment Configuration and Measurement Protocol}\label{app:mobile_setup}

We export the trained models to native mobile inference pipelines on two specific phones: an iPhone 17 Pro Max with Apple A19 Pro, and an Android mobile phone with MediaTek Dimensity 9400. The iOS implementation uses an MLX Swift application that loads an exported MLX bundle, while the Android implementation uses a Kotlin/JNI application backed by a native arm64-v8a C++ runtime with an optimized packed \texttt{complex\_phase\_v2} wide-linear backend. These measurements therefore reflect our native deployment implementations rather than Core ML or NNAPI baselines. For both platforms, the FP32 baseline and the RPFQ model use the same image preprocessing, input resolution, batch size, and application stack on the corresponding device.

All measurements use batch size 1 and \(224 \times 224\) inputs. We exclude camera capture, image decoding, network transfer, and UI rendering, and time only the inference path from a preprocessed input tensor to the Top-1 prediction. Each run performs 50 warm-up iterations and 500 timed iterations. We report median latency (p50), tail latency (p90), and model loading time so that model-initialization effects are not conflated with steady-state inference. Model size is measured as the exported on-disk container rather than a ZIP-compressed archive. For RPFQ, the container stores 2-bit packed weight codes together with group-wise scales, rotations, anchors, biases, and the remaining non-quantized modules. Concretely, in the default \(K=4\), \(W2\) setting, each group stores 7 metadata scalars: one rotation \(\theta_g\), four anchors \(\{\phi_k\}_{k=1}^{4}\), and two scales \((\alpha_g,\beta_g)\). Thus the metadata cost is exactly 7 scalars per group, i.e., \(7b_{\text{meta}}\) bits per group or \(7b_{\text{meta}}/(2G)\) bits per original real weight for a group of \(G\) complex weights. The mobile exporters preserve the packed 2-bit weight representation but do not export a native INT4 activation path: the Apple bundle stores packed phase-coded linear weights plus floating-point tensors for the remaining modules, and the Android bundle manifest explicitly sets \texttt{activation\_quantization=false}. Accordingly, the \(W/A\) notation in Table~\ref{tab:mobile} identifies the training/evaluation quantization setting of the source checkpoint rather than an end-to-end on-phone integer activation pipeline. Inference on both phones therefore runs packed low-bit weights together with floating-point activations, LayerNorm, GELU, Softmax, residual additions, patch embedding, and classification logic. After export, the treatment of packed weights is runtime-dependent: the backend may keep packed storage, unpack weights ahead of execution, or partially convert them during kernel dispatch. We therefore report end-to-end latency, load time, and model size rather than a stagewise kernel decomposition. Our current harness also does not expose stable, directly comparable per-inference energy counters across the two mobile stacks, so we leave power and energy measurements to future work.

\subsection{Training-Efficiency Profiling Details}\label{app:train_efficiency}

All training-efficiency numbers are obtained on a single node with 8\(\times\)NVIDIA H800-80GB GPUs, batch size 64 per GPU (global batch 512), and 2{,}502 iterations per epoch on ImageNet-1K. RPFQ statistics are parsed directly from raw training logs; we report the mean iteration time after warm-up and project the 300-epoch wall-clock from it. For Q-ViT we run the official codebase\footnote{\url{https://github.com/YanjingLi0202/Q-ViT}.} under the same global batch size and profile the first \(\sim\!500\) iterations after warm-up. For OFQ\footnote{\url{https://github.com/nbasyl/OFQ}.}, the official repository provides training scripts only for DeiT-T/-S and Swin-T, and the original OFQ paper does not report ViT-B/16 runtime~\citep{liu2023ofq}; we therefore report only the schedule-level lower bound implied by its official two-stage protocol. This interpretation is consistent with recent third-party reports that distillation-heavy ViT QAT methods incur high training cost and memory pressure~\citep{liu2025gplq}.

\begin{table}[t]
  \caption{Training efficiency on ImageNet-1K with 8\(\times\)H800-80GB. RPFQ numbers are parsed from full training logs. Q-ViT runtimes are profiled from its official codebase under the same global batch size and projected to 300 epochs. For OFQ, we report only the schedule-level lower bound implied by its public two-stage protocol because no ViT-B/16 runtime is available.}
  \label{tab:train_efficiency}
  \centering
  \small
  \setlength{\tabcolsep}{4.4pt}
  \renewcommand{\arraystretch}{1.10}
  \resizebox{\linewidth}{!}{%
  \begin{tabular}{llcccc}
    \toprule
    Method & Backbone / setting & Protocol & Aux.\ branch
           & Iter (s) & 300-epoch total \\
    \midrule
    \oursrow
    \oursmethod & ViT-B/16 W2/A4 & single-stage QAT & none
                & \textbf{0.223} & \textbf{46.55\,h / 1.94\,d} \\
    \oursrow
    \oursmethod & DeiT-S W2/A8 & single-stage QAT & none
                & 0.191 & 39.75\,h / 1.66\,d \\
    \oursrow
    \oursmethod & Swin-T W2/A8 & single-stage QAT & none
                & 0.216 & 45.06\,h / 1.88\,d \\
    \midrule
    Q-ViT & ViT-B/16 W2/A4 & single-stage QAT
          & teacher + DGD attention distill.
          & 0.768 & 160.21\,h / 6.68\,d \\
    OFQ   & DeiT-S / Swin-T (no ViT-B recipe) & QAT + CGA fine-tune
          & teacher + QKR + CGA
          & --    & \(\geq 600\) epochs \\
    \bottomrule
  \end{tabular}}
\end{table}

On ViT-B/16, RPFQ finishes a 300-epoch run in 46.55 hours, compared with 160.21 projected hours for Q-ViT under the same global batch size, a 3.44\(\times\) reduction in wall-clock time. OFQ's public recipe already implies at least 600 training epochs, i.e., at least 2\(\times\) the schedule length of RPFQ before accounting for per-step overhead. The memory story is similar: Q-ViT reaches a logged peak of 61{,}979 MiB, whereas the DGD auxiliary tensors alone contribute an analytical lower bound of 7.06 GiB, excluding teacher activations and autograd buffers. By contrast, RPFQ adds only a small set of per-group angles, anchors, and scales, whose total size is negligible relative to the model. Overall, RPFQ is not only accurate at 2-bit weights, but also substantially cheaper to train than distillation- or reparameterization-heavy ViT QAT baselines~\citep{liu2025gplq}.

For the memory comparison, the dominant extra cost of Q-ViT comes from the FP32 teacher and the DGD Q/K similarity tensors. With local batch \(B{=}64\), token length \(N{=}197\), \(L{=}12\) layers, \(H{=}12\) heads, and head dimension \(d_h{=}64\), an analytical lower bound for the DGD auxiliary tensors is
\[
  \underbrace{4\,B L H N^{2} \cdot 4\,\text{B}}_{\text{Q/K Gram, both branches}}
  \;+\;
  \underbrace{4\,B L H N d_h \cdot 4\,\text{B}}_{\text{Q/K features, both branches}}
  \;\approx\; 7.06\,\text{GiB},
\]
where the factor \(4\) accounts for student-Q, student-K, teacher-Q, and teacher-K. This lower bound excludes teacher activations, optimizer states, autograd temporaries, and CUDA workspace, so the realized overhead is strictly larger.

\newpage
\input{checklist.tex}

\end{document}

%% file: checklist.tex
\section*{NeurIPS Paper Checklist}

\begin{enumerate}

\item {\bf Claims}
    \item[] Question: Do the main claims made in the abstract and introduction accurately reflect the paper's contributions and scope?
    \item[] Answer: \answerYes{} % Replace by \answerYes{}, \answerNo{}, or \answerNA{}.
    \item[] Justification: The abstract and Section~1 state the problem scope, the phase-geometry motivation, the proposed quantizer, and the empirical/mobile claims, and these are supported by the results in Section~4.

\item {\bf Limitations}
    \item[] Question: Does the paper discuss the limitations of the work performed by the authors?
    \item[] Answer: \answerYes{} % Replace by \answerYes{}, \answerNo{}, or \answerNA{}.
    \item[] Justification: The ``Discussion and Limitations'' section discusses the restricted evaluation scope, the practical-vs.-theoretical gap in mobile speedups, the absence of custom low-bit kernel attribution or energy measurements in the current deployment study, and the remaining instability of A2 activation quantization.

\item {\bf Theory assumptions and proofs}
    \item[] Question: For each theoretical result, does the paper provide the full set of assumptions and a complete (and correct) proof?
    \item[] Answer: \answerNA{} % Replace by \answerYes{}, \answerNo{}, or \answerNA{}.
    \item[] Justification: The paper is primarily methodological and empirical rather than theorem-driven. The paired-complex reparameterization and least-squares scale formula are direct constructions described in Section~3 and Appendix~\ref{app:paired}, not standalone theorem/proof claims.

    \item {\bf Experimental result reproducibility}
    \item[] Question: Does the paper fully disclose all the information needed to reproduce the main experimental results of the paper to the extent that it affects the main claims and/or conclusions of the paper (regardless of whether the code and data are provided or not)?
    \item[] Answer: \answerYes{} % Replace by \answerYes{}, \answerNo{}, or \answerNA{}.
    \item[] Justification: Sections~4.1--4.6 and Appendices~\ref{app:exp_setup}--\ref{app:train_efficiency} disclose the training recipe, evaluation protocol, ablations, deployment protocol, and profiling assumptions needed to reproduce the main results.

\item {\bf Open access to data and code}
    \item[] Question: Does the paper provide open access to the data and code, with sufficient instructions to faithfully reproduce the main experimental results, as described in supplemental material?
    \item[] Answer: \answerYes{} % Replace by \answerYes{}, \answerNo{}, or \answerNA{}.
    \item[] Justification: The experiments use the public ImageNet-1K benchmark, and the supplementary ZIP / anonymous repository include anonymized code, configuration files, reproduction scripts, and a README with environment and launch instructions; see Appendix~\ref{app:exp_setup}.

\item {\bf Experimental setting/details}
    \item[] Question: Does the paper specify all the training and test details (e.g., data splits, hyperparameters, how they were chosen, type of optimizer) necessary to understand the results?
    \item[] Answer: \answerYes{} % Replace by \answerYes{}, \answerNo{}, or \answerNA{}.
    \item[] Justification: Section~4.1 specifies datasets, models, quantization scope, baselines, and evaluation protocol, including the matched 300-epoch recipe, the shared final-checkpoint evaluation rule for FP32 and quantized runs trained in our codebase, and the three-seed averaging protocol for the non-reported main-table results. Appendices~\ref{app:exp_setup}--\ref{app:pairing_ablation} and~\ref{app:mobile_setup} provide the training, reproduction, pairing-ablation, and deployment details.

\item {\bf Experiment statistical significance}
    \item[] Question: Does the paper report error bars suitably and correctly defined or other appropriate information about the statistical significance of the experiments?
    \item[] Answer: \answerYes{} % Replace by \answerYes{}, \answerNo{}, or \answerNA{}.
    \item[] Justification: The non-reported main-table results are reported as means over three independent runs with different seeds, and Appendix~\ref{app:stat_uncert} additionally reports conservative 95\% confidence intervals for the main Top-1 comparisons on the fixed 50{,}000-image ImageNet-1K validation set. That appendix states the variability source being quantified (finite-sample validation uncertainty), gives the closed-form binomial approximation used for the intervals, and reports confidence intervals for the main RPFQ-vs-baseline gaps.

\item {\bf Experiments compute resources}
    \item[] Question: For each experiment, does the paper provide sufficient information on the computer resources (type of compute workers, memory, time of execution) needed to reproduce the experiments?
    \item[] Answer: \answerYes{} % Replace by \answerYes{}, \answerNo{}, or \answerNA{}.
    \item[] Justification: Section~4.4 and Appendix~\ref{app:train_efficiency} report GPU type, memory, batch size, iteration time, wall-clock projections, and memory analysis for training, and Appendix~\ref{app:mobile_setup} reports the mobile runtime stack, hardware class, warm-up/timing protocol, load-time measurement, and deployment-scope caveats for the phone experiments.
    
\item {\bf Code of ethics}
    \item[] Question: Does the research conducted in the paper conform, in every respect, with the NeurIPS Code of Ethics \url{https://neurips.cc/public/EthicsGuidelines}?
    \item[] Answer: \answerYes{} % Replace by \answerYes{}, \answerNo{}, or \answerNA{}.
    \item[] Justification: The work uses public benchmark data and standard model training, does not involve human subjects, and discusses limitations and deployment risks in the paper. We are not aware of any aspect of the study that conflicts with the NeurIPS Code of Ethics.

\item {\bf Broader impacts}
    \item[] Question: Does the paper discuss both potential positive societal impacts and negative societal impacts of the work performed?
    \item[] Answer: \answerYes{} % Replace by \answerYes{}, \answerNo{}, or \answerNA{}.
    \item[] Justification: The paper discusses both positive impacts (lower storage, memory traffic, latency, and potentially energy use on resource-constrained devices) and negative impacts (easier deployment of vision systems in surveillance or other sensitive settings without sufficient task-specific evaluation) in the ``Discussion and Limitations section.
    
\item {\bf Safeguards}
    \item[] Question: Does the paper describe safeguards that have been put in place for responsible release of data or models that have a high risk for misuse (e.g., pre-trained language models, image generators, or scraped datasets)?
    \item[] Answer: \answerNA{} % Replace by \answerYes{}, \answerNo{}, or \answerNA{}.
    \item[] Justification: The paper does not release high-risk generative models, scraped datasets, or other assets that typically require special access controls. It studies compression of standard ImageNet classification models and therefore does not call for additional release safeguards beyond the cautions discussed in the paper.

\item {\bf Licenses for existing assets}
    \item[] Question: Are the creators or original owners of assets (e.g., code, data, models), used in the paper, properly credited and are the license and terms of use explicitly mentioned and properly respected?
    \item[] Answer: \answerYes{} % Replace by \answerYes{}, \answerNo{}, or \answerNA{}.
    \item[] Justification: Appendix~\ref{app:assets} explicitly credits the main external assets used in the paper and summarizes their licenses or access terms, including ImageNet-1K access restrictions, the Apache-2.0 license for \texttt{timm}, the MIT license for OFQ, and our non-redistributive internal use of the public Q-ViT repository.

\item {\bf New assets}
    \item[] Question: Are new assets introduced in the paper well documented and is the documentation provided alongside the assets?
    \item[] Answer: \answerYes{} % Replace by \answerYes{}, \answerNo{}, or \answerNA{}.
    \item[] Justification: The supplementary materials include anonymized source code, configuration files, reproduction scripts, and a README with setup and launch instructions; see Appendix~\ref{app:exp_setup}. These materials document the released implementation sufficiently for reproduction.

\item {\bf Crowdsourcing and research with human subjects}
    \item[] Question: For crowdsourcing experiments and research with human subjects, does the paper include the full text of instructions given to participants and screenshots, if applicable, as well as details about compensation (if any)? 
    \item[] Answer: \answerNA{} % Replace by \answerYes{}, \answerNo{}, or \answerNA{}.
    \item[] Justification: The paper does not involve crowdsourcing or research with human subjects.

\item {\bf Institutional review board (IRB) approvals or equivalent for research with human subjects}
    \item[] Question: Does the paper describe potential risks incurred by study participants, whether such risks were disclosed to the subjects, and whether Institutional Review Board (IRB) approvals (or an equivalent approval/review based on the requirements of your country or institution) were obtained?
    \item[] Answer: \answerNA{} % Replace by \answerYes{}, \answerNo{}, or \answerNA{}.
    \item[] Justification: The paper does not involve crowdsourcing or research with human subjects, so IRB approval and participant-risk disclosures are not applicable.

\item {\bf Declaration of LLM usage}
    \item[] Question: Does the paper describe the usage of LLMs if it is an important, original, or non-standard component of the core methods in this research? Note that if the LLM is used only for writing, editing, or formatting purposes and does \emph{not} impact the core methodology, scientific rigor, or originality of the research, declaration is not required.
    %this research? 
    \item[] Answer: \answerNA{} % Replace by \answerYes{}, \answerNo{}, or \answerNA{}.
    \item[] Justification: LLMs are not part of the proposed method, experimental pipeline, or scientific claims. Any writing or editing assistance does not affect the core methodology or originality of the research.

\end{enumerate}